\documentclass[10pt,twocolumn,letterpaper]{article}

\usepackage{cvpr}      
\definecolor{cvprblue}{rgb}{0.21,0.49,0.74}
\usepackage[pagebackref,breaklinks,colorlinks,allcolors=cvprblue]{hyperref}
\usepackage[linesnumbered,ruled,vlined]{algorithm2e}
\usepackage{booktabs}
\usepackage{bm}
\usepackage{amsmath}
\usepackage{graphicx}
\usepackage{stfloats}
\usepackage{caption}
\usepackage{tikz}
\usepackage{indentfirst}
\usepackage{multirow}
\usepackage{hyperref}
\usepackage{longtable}
\usepackage[pass]{geometry}
\usepackage{float}

\def\paperID{9698} 
\def\confName{CVPR}
\def\confYear{2026}

\title{\textsc{SceneFoundry}: Generating Interactive Infinite 3D Worlds}

\author{
  ChunTeng Chen$^{1}$ \quad YiChen Hsu$^{2}$ \quad YiWen Liu$^{1}$ \quad WeiFang Sun$^{3}$ \\
  TsaiChing Ni$^{1}$ \quad ChunYi Lee$^{4}$ \quad Min Sun$^{2}$ \quad YuanFu Yang$^{1}$ \\[2mm]
  $^{1}$National Yang Ming Chiao Tung University \quad $^{2}$National Tsing Hua University \\
  $^{3}$NVIDIA AI Technology Center \quad $^{4}$National Taiwan University
}

\begin{document}
\twocolumn[{%
\renewcommand\twocolumn[1][]{#1}
\maketitle
\begin{center}
    \centering
    \includegraphics[width=1.0\linewidth]{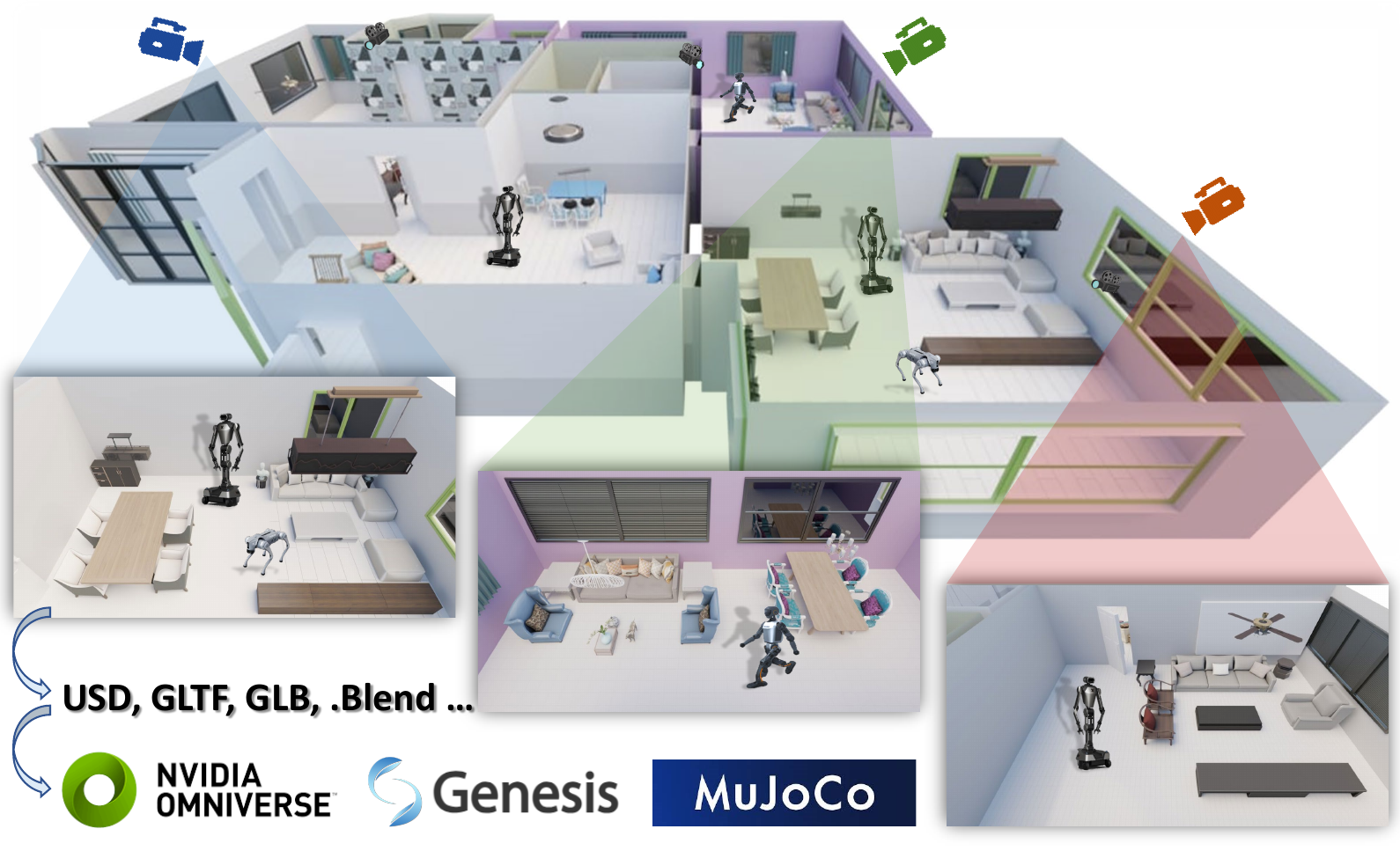}
    \vspace{-0.5em}
    \captionsetup{hypcap=false}
    \captionof{figure}{
        Overview of \textbf{SceneFoundry}. The framework generates apartment-scale 3D scenes from natural language prompts via LLM-guided floor plan generation, diffusion-based placement, and post-optimization ensuring articulated functionality and robot navigability.
    }
    \label{fig:teaser}
\end{center}
}]

\maketitle
\begin{abstract}
\indent The ability to automatically generate large-scale, interactive, and physically realistic 3D environments is crucial for advancing robotic learning and embodied intelligence. However, existing generative approaches often fail to capture the functional complexity of real-world interiors, particularly those containing articulated objects with movable parts essential for manipulation and navigation. This paper presents SceneFoundry, a language-guided diffusion framework that generates apartment-scale 3D worlds with functionally articulated furniture and semantically diverse layouts for robotic training. From natural language prompts, an LLM module controls floor layout generation, while diffusion-based posterior sampling efficiently populates the scene with articulated assets from large-scale 3D repositories. To ensure physical usability, SceneFoundry employs differentiable guidance functions to regulate object quantity, prevent articulation collisions, and maintain sufficient walkable space for robotic navigation. Extensive experiments demonstrate that our framework generates structurally valid, semantically coherent, and functionally interactive environments across diverse scene types and conditions, enabling scalable embodied AI research.
\end{abstract}
\vspace{-0.5cm}
\section{Introduction}
\label{sec:intro}
The ability to generate diverse, large-scale, and realistic 3D indoor environments is fundamental to the advancement of robotics \cite{savva2019habitatplatformembodiedai}, virtual reality, and embodied AI \cite{kolve2022ai2thorinteractive3denvironment}. However, bridging the simulation-to-reality gap remains a significant hurdle \cite{chen2023genaugretargetingbehaviorsunseen}, often because generated simulation environments lack the complexity, controllability, and functional realism of their physical counterparts.

Recent efforts \cite{yang2024physcenephysicallyinteractable3d, huang2023diffusionbasedgenerationoptimizationplanning, liang2025sinfrealisticindoorscene} have focused on increasing the physical realism and interactability of simulated environments. While enhancing functional realism in specific aspects, such approaches can inadvertently compromise the visual realism or diversity of the layouts. A major limitation of many learning-based methods is their inability to generate complete apartment-scale layouts, as they often focus only on single rooms. Procedural generation frameworks like Infinigen \cite{raistrick2024infinigenindoorsphotorealisticindoor} can produce such large-scale environments, but they are computationally intensive. Existing works \cite{yang2024physcenephysicallyinteractable3d, tang2024diffuscenedenoisingdiffusionmodels, paschalidou2021atissautoregressivetransformersindoor} also often lack fine-grained control over crucial scene properties, which is essential for generating targeted training data distributions.

This paper introduces a multi-stage, controllable generative framework designed to generate apartment-scale 3D indoor scenes that are not only visually diverse but also semantically coherent and functionally sound. Our approach bridges the gap between high-level user intent, specified via natural language, and the generation of structurally valid layouts suitable for robotic interaction. We architect a pipeline that integrates semantic guidance with a suite of novel constraints to ensure that every generated scene is tailored to specific requirements and is physically usable. As shown in Figure~\ref{fig:teaser}, SceneFoundry generates photorealistic, apartment-scale, and controllable 3D scenes. Our contributions are summarized below.

\begin{itemize}
    \item \textbf{LLM-based Parameter Space Guidance.} We introduce a module that translates abstract user commands into low-level parameters, enabling semantic control over the generative priors of a floor plan generator.

    \item \textbf{Novel Functional Guidance Mechanisms.} 
    We introduce a set of differentiable constraints to enforce functional plausibility. This includes:
    \begin{itemize}
        \item An \textbf{Object Quantity Control} for precisely enforcing the number of objects in a scene.
        \item An \textbf{Articulated Object Collision Constraint} that penalizes configurations where functional parts are obstructed, ensuring interactability.
    \end{itemize}

    \item \textbf{Walkable Area Control.} A final Walkable Area Control optimization is applied to the generated layout to refine spatial density and guarantee agent navigability.

    \item \textbf{Novel Evaluation Metrics.} To validate the effectiveness of the generation and control methods, we introduce new evaluation metrics that measure controllability.
\end{itemize}
\section{Related Work}
\label{sec:related_work}

\textbf{Indoor Scene Layout Generation.}
The automated generation of 3D indoor scenes is a long-standing challenge in computer graphics and vision. Early approaches relied on procedural generation, employing rule-based grammars or optimization techniques to synthesize layouts. A prominent recent example, Infinigen \cite{raistrick2024infinigenindoorsphotorealisticindoor}, utilizes procedural methods combined with simulated annealing to generate high-fidelity, apartment-scale layouts. While capable of producing complex and realistic results, these methods are often computationally intensive, time-consuming, and difficult to control without expert knowledge of the underlying rules.

Learning-based approaches are now dominant. Autoregressive models, such as ATISS \cite{paschalidou2021atissautoregressivetransformersindoor}, generate objects sequentially. While this models inter-object relationships, it suffers from error accumulation, slow sampling, and difficult holistic editing. Diffusion models \cite{ho2020denoisingdiffusionprobabilisticmodels} are a powerful alternative. Methods like DiffuScene \cite{tang2024diffuscenedenoisingdiffusionmodels} generate all object parameters in parallel, offering superior holistic coherence, editing flexibility, and state-of-the-art quality \cite{dhariwal2021diffusionmodelsbeatgans}. We therefore adopt this paradigm.

Effective generative modeling critically depends on both dataset quality and structural consistency. While large-scale datasets of real-world scans like ScanNet \cite{dai2017scannetrichlyannotated3dreconstructions} and Matterport3D \cite{chang2017matterport3dlearningrgbddata} are invaluable for reconstruction and navigation, they are less suitable for \textit{generative} tasks. We therefore utilize clean, CAD-based datasets, 3D-FRONT \cite{fu20213dfront3dfurnishedrooms} and GAPartNet \cite{geng2023gapartnetcrosscategorydomaingeneralizableobject}, which offer well-structured geometry and part-level semantics ideal for controllable 3D scene synthesis.

\textbf{Guidance of Diffusion Model.}
Controlling generative models is crucial. Early studies introduced classifier guidance\cite{ho2022classifierfreediffusionguidance}, which leverages the gradient of an external classifier to steer the sampling process. This gradient-based steering concept was later generalized. The core idea, often referred to as diffusion posterior sampling, is highly flexible and allows for guidance through any differentiable function, not just a classifier. This principle is ideal for enforcing functional 3D constraints, and recent work has begun to explore its use for physical plausibility or robot reachability\cite{yang2024physcenephysicallyinteractable3d, bansal2023universalguidancediffusionmodels, ho2022classifierfreediffusionguidance}. We adopt this posterior sampling approach for its modularity, training a single unconditional model and applying diverse constraints at test time while avoiding the cost of multiple specialized conditional models.

\begin{figure*}[t!]
    \centering
    \includegraphics[width=1.0\linewidth]{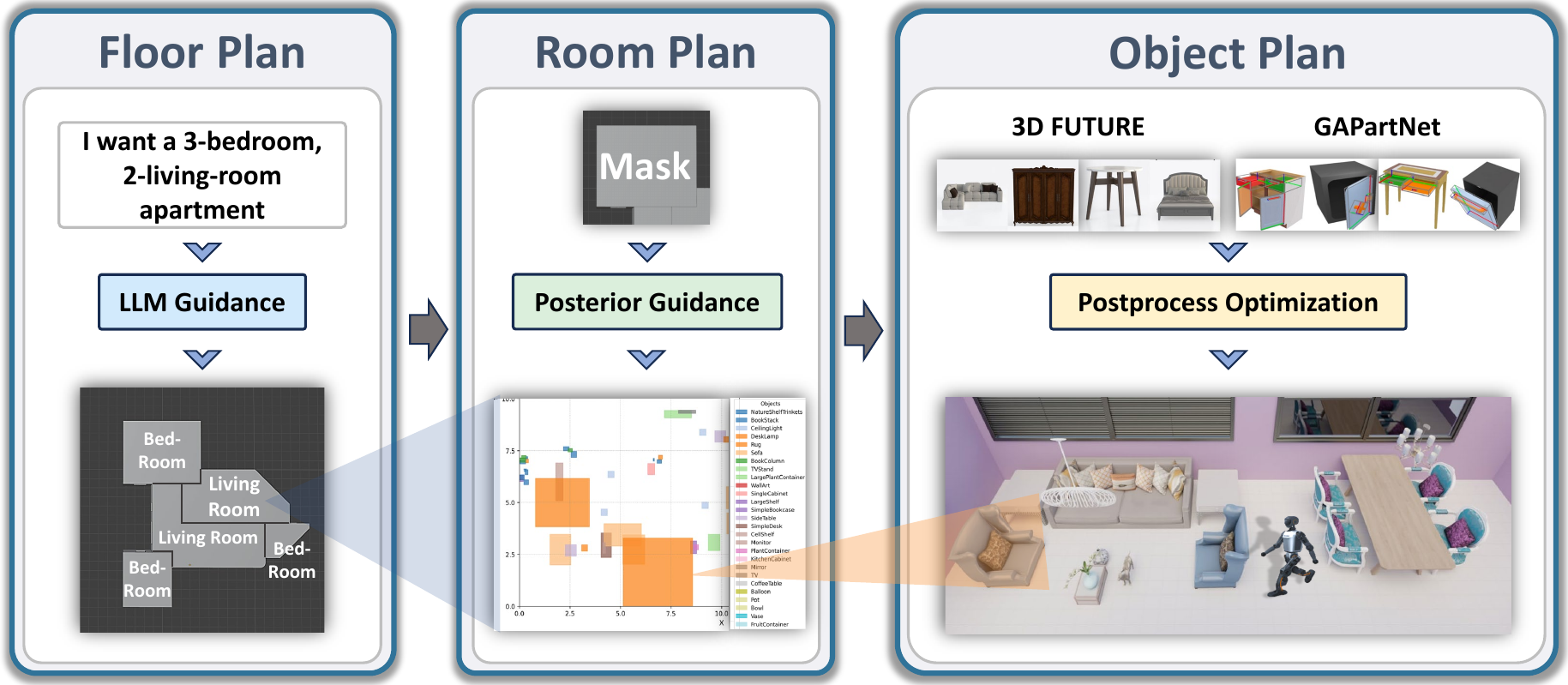}
    \caption{Overview of our apartment-scale generation pipeline. An LLM first guides procedural floor plan generation (Sec.~\ref{sec:llm_guide}), diffusion posterior guidance generates plausible room bounding boxes (Sec.~\ref{sec:posterior_sampling}, Sec.~\ref{sec:control_quantity}, Sec.~\ref{sec:control_articoll}), and 3D assets from 3D-FRONT/GAPartNet are refined via post-optimization to complete the layout (Sec.~\ref{sec:control_walkable}).}
    \label{fig:pipeline}
\end{figure*}

\begin{figure}[b]
    \centering
    \includegraphics[width=1.0\linewidth]{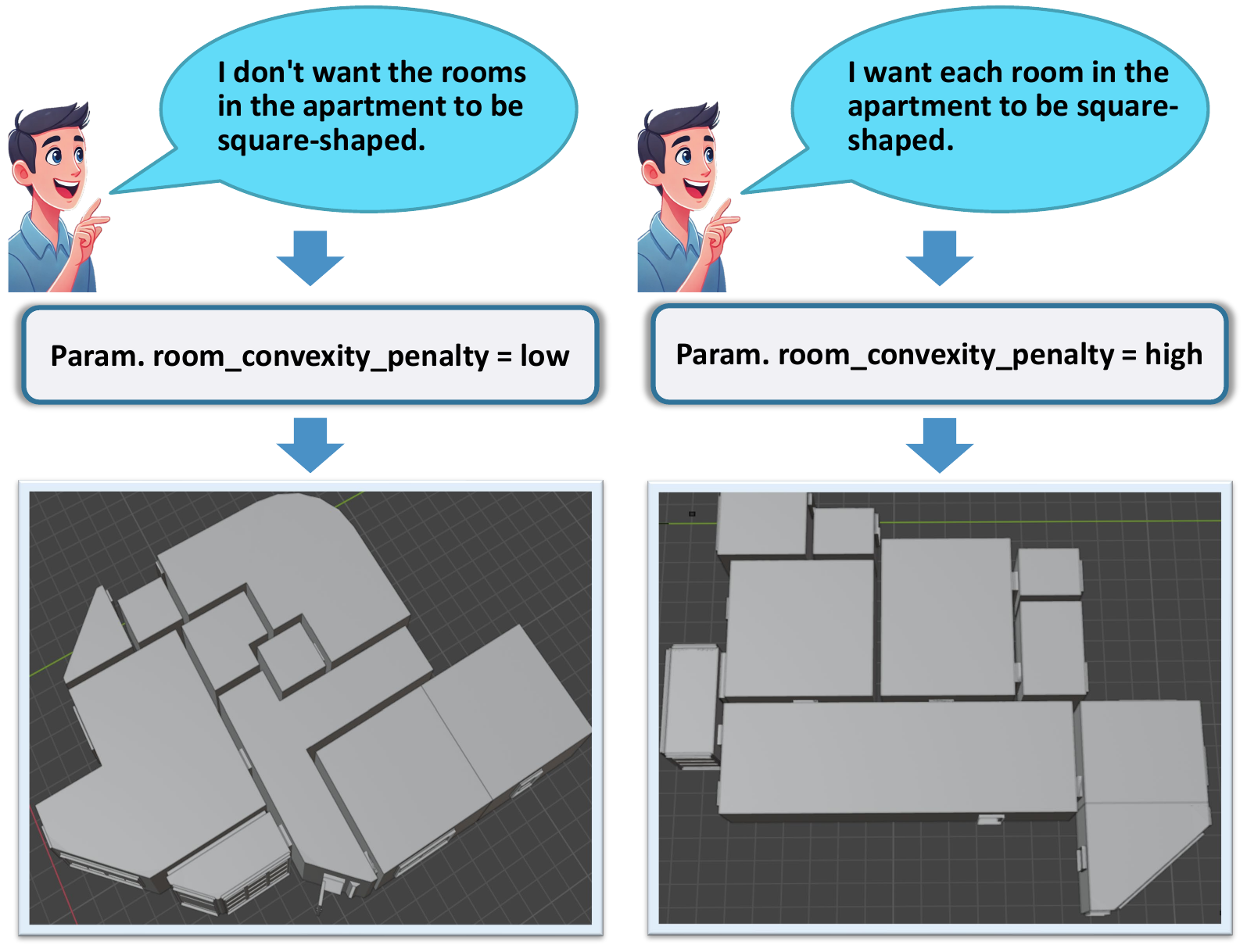}
    \caption{Illustration of our LLM-based Guidance. A low penalty (left) produces diverse, non-rectilinear layouts, whereas a high penalty (right) enforces square-shaped room layouts.}
    \label{fig:llm_guidance}
\end{figure}

\section{SceneFoundry}
\label{sec:method}
Our framework adopts a multi-stage pipeline to generate controllable, apartment-scale 3D scenes for robot training, as illustrated in Figure~\ref{fig:pipeline}. The pipeline begins with an LLM-based parameter space generation (Sec.~\ref{sec:llm_guide}) that translates user prompts into low-level parameters for floor plan generation. A diffusion model employing posterior sampling (Sec.~\ref{sec:posterior_sampling}) then populates these layouts with furniture assets.

To ensure functional viability, we integrate three control mechanisms. During sampling, the model is guided by differentiable guidance functions: Object Quantity Control (Sec.~\ref{sec:control_quantity}) and the proposed Articulated Object Collision Constraint (Sec.~\ref{sec:control_articoll}) to maintain usability of movable parts. A Walkable Area Control post-processing step (Sec.~\ref{sec:control_walkable}) further refines spatial density to guarantee navigability. We also introduce a set of evaluation metrics(Sec~\ref{sec:evaluation metric 3}) to quantitatively evaluate the effectiveness of our methods.

\begin{figure*}[t!]
    \centering
    \includegraphics[width=1.0\linewidth]{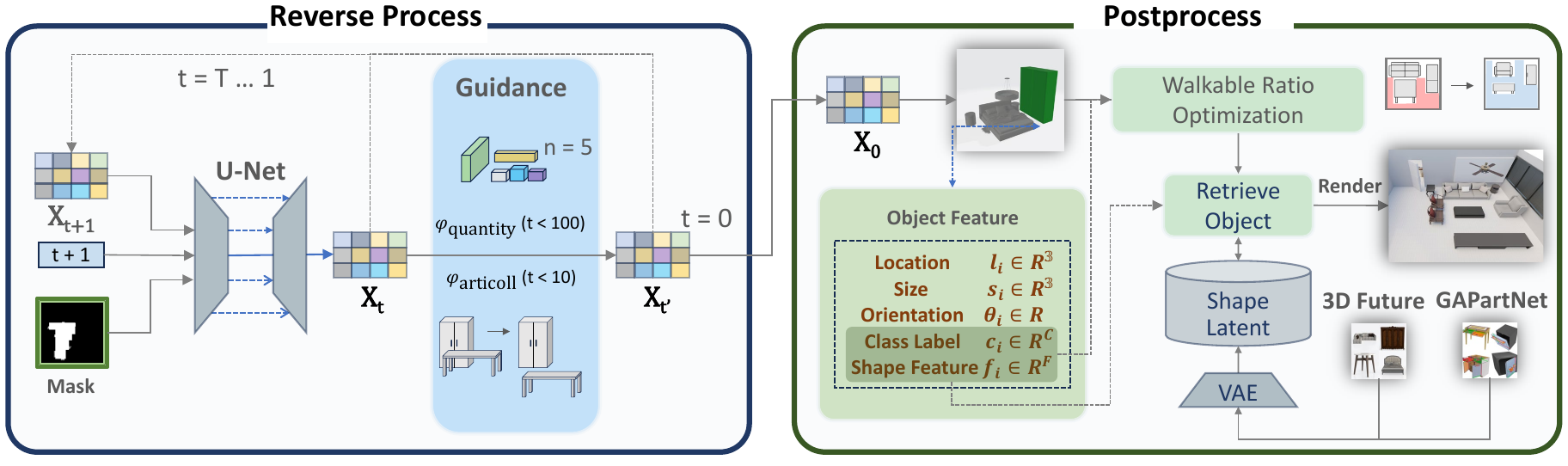}
    \caption{Guidance scheduling during the reverse diffusion process. Object quantity control is applied at $t < 100$ and articulated collision constraint at $t < 1$0, followed by a final walkable-ratio optimization at $t = 0$ to generate a realistic scene.
    }
    \label{fig:model}
\end{figure*}

\subsection{LLM-Guided Parameter Space Generation}
\label{sec:llm_guide}
We adopt the Infinigen~\cite{raistrick2024infinigenindoorsphotorealisticindoor} framework, which generates room layouts through a simulated annealing process governed by twelve reward functions. However, its high-dimensional parameter space is not intuitive to control for users. To enhance usability, we design an LLM-based parameterization framework that interprets natural-language prompts and produces corresponding parameters for these functions, as shown in Figure~\ref{fig:llm_guidance}. This semantic-to-parameter mapping converts abstract user descriptions into concrete floor plans while preserving the stochastic diversity of Infinigen.

\subsection{Diffusion Posterior Sampling}
\label{sec:posterior_sampling}
Our method steers the reverse diffusion trajectory using a composite guidance function, $\varphi(\cdot)$, which enforces explicit constraints on the generated 3D scenes. This approach adapts the principles of diffusion posterior sampling to ensure structural validity in the generated layouts, as shown in the reverse process part of Figure~\ref{fig:model}.

\textbf{Object Feature.}
Following \cite{tang2024diffuscenedenoisingdiffusionmodels}, we represent a 3D scene $\mathbf{x}$ as an unordered set of $N$ objects, $\{\mathbf{o}_i\}^N_{i=1}$. Each object $\mathbf{o}_i$ is a vector $[ \mathbf{l}_i, \mathbf{s}_i, \mathbf{\theta}_i, \mathbf{c}_i, \mathbf{f}_i]$ encoding its location, size, orientation, semantics, and a latent shape feature. This latent space is derived from a pre-trained VAE, following \cite{tang2024diffuscenedenoisingdiffusionmodels, yang2024physcenephysicallyinteractable3d}.The generated latent feature $\mathbf{f}_i$ is used for a nearest-neighbor search to retrieve the best-matching asset from either 3D-FRONT or GAPartNet, as shown in postprocess part of Figure~\ref{fig:model}. This allows us to compose novel scenes that cohesively integrate both static and articulated objects.

\textbf{Training.}
We adopt a \textit{constraint-guided learning} strategy. Instead of a standard denoising objective, the model $\bm{\epsilon}_{\theta}$ is trained to predict the noise $\bm{\epsilon}$ while simultaneously anticipating the constraint gradient $\mathbf{g}$ (from our constraint functions $\varphi$). This is optimized by minimizing a guided $\mathcal{L}_2$ loss:
\begin{equation}
\label{eq:denoising_loss}
\mathcal{L} = \mathbb{E}_{t, \mathbf{x}_0, \bm{\epsilon}} \left[ \| (\bm{\epsilon} - \lambda\mathbf{\Sigma}\mathbf{g}) - \bm{\epsilon}_{\theta}(\mathbf{x}_t, t, \mathcal{F}) \|^2_2 \right]
\end{equation}
This approach embeds knowledge of the constraints directly into the model weights during the training phase.

\textbf{Sampling.}
During inference, we perform an iterative reverse process starting from $\mathbf{x}_T \sim \mathcal{N}(\mathbf{0}, \mathbf{I})$. At each step $t$, the model first predicts the parameters ($\bm{\mu}_{\theta}$, $\mathbf{\Sigma}_{\theta}$) of the unguided posterior $p_{\theta}(\mathbf{x}_{t-1}|\mathbf{x}_t)$. We then compute the gradient of our composite guidance function, $\nabla_{\mathbf{x}_t}\varphi(\mathbf{x}_t)$, and use it to perturb the predicted mean, steering the sampling step towards valid regions:
{\small
\begin{equation}
\label{eq:guided_step}
\begin{split}
\mathbf{x}_{t-1} \sim \mathcal{N}\Big(& \bm{\mu}_{\theta}(\mathbf{x}_t, t, \mathcal{F}) \\
& + \lambda\mathbf{\Sigma}_{\theta}(\mathbf{x}_t, t, \mathcal{F})\nabla_{\mathbf{x}_t}\varphi(\mathbf{x}_t, \mathcal{F}), \mathbf{\Sigma}_{\theta}(\mathbf{x}_t, t, \mathcal{F}) \Big)
\end{split}
\end{equation}
}
Iterating this process yields the final sample $\mathbf{x}_0$. The complete procedure is shown in Algorithm~\ref{alg:diffuser_sample}.

\begin{algorithm}[t!]
    \SetAlgoLined
    \caption{\small Guidance Sampling in Model}
    \label{alg:diffuser_sample}
    \SetKwInOut{module}{Modules}
    \SetKwProg{Fn}{function}{:}{}
    \SetKwFunction{sample}{{\bf sample}}
    \module{Model $p_{\theta}(\cdot | \mathcal{F})$, guidance functions $\varphi(\cdot)=\{\varphi_{\text{quantity}}(\cdot), \varphi_{\text{articoll}}(\cdot)\}$.}
    
    \texttt{// constraint-guided learning}\\
    \KwIn{3D scene layout $\mathbf{x} = \{\mathbf{o}_1, \dots, \mathbf{o}_N\}$ and floor plan $\mathcal{F}$, where N is a fixed number of objects.}
    \Repeat{\text{converged}}{    
        $\mathbf{x}_0 \sim p(\mathbf{x}_0 | \mathcal{F})$ \\
        $\bm{\epsilon} \sim \mathcal{N}(\mathbf{0}, \mathbf{I})$, $t \sim \mathcal{U}(\{1,\dots,T\})$\\
        $\mathbf{x}_t = \sqrt{\hat{\alpha}_t}\mathbf{x}_0 + \sqrt{1 - \hat{\alpha}_t}\bm{\epsilon}$, $\tilde{\mathbf{x}}_0^t \sim p_{\theta}(\cdot)$\\
        $\theta = \theta - \eta\nabla_\theta \| \bm{\epsilon} - \bm{\epsilon}_{\theta}(\mathbf{x}_t,t) - \lambda\mathbf{\Sigma}\mathbf{g}\|^2_2$
    }

    \texttt{// one-step guided sampling}\\
    \Fn{\sample$(\bm{\tau}^{t}$, $\varphi$)}{
        $\bm{\mu} = \bm{\mu}_{\theta}(\mathbf{x}_t,t, \mathcal{F})$, $\mathbf{\Sigma} = \mathbf{\Sigma}_{\theta}(\mathbf{x}_t,t, \mathcal{F})$\\
        $\varphi(\mathbf{x}_t) = \gamma_1\varphi_{\text{quantity}}(\mathbf{x}_t)+\gamma_2\varphi_{\text{articoll}}(\mathbf{x}_t)$\\
        $\mathbf{x}_{t-1} = \mathcal{N}(\mathbf{x}_{t-1}; \bm{\mu} + \lambda\mathbf{\Sigma}\nabla_{\mathbf{x}_t}\varphi(\mathbf{x}_t, \mathcal{F})|_{\mathbf{x}_t=\bm{\mu}}, \mathbf{\Sigma})$\\
        \Return $\mathbf{x}_{t-1}$ \\
    }
    
    \texttt{// constraint-guided generation}\\
    \KwIn{initial scene layout $\mathbf{x}_T \sim \mathcal{N}(\mathbf{0}, \mathbf{I})$}
    \For{$t=T,\dots,1$}{
        \texttt{// sampling with optimization}\\
        $\mathbf{x}_{t-1}=\sample(\mathbf{x}_{t}, \varphi)$\\
    }
    \Return $\mathbf{x}_0$ \\
\end{algorithm}

\subsection{Object Quantity Constraint}
\label{sec:control_quantity}

To control object quantity, we introduce a differentiable guidance function, $\varphi_{\text{quantity}}$, which operates on the predicted class logits during the reverse diffusion process. Our scene representation uses $N_{max}$ potential object slots, where each slot's logits include a channel for an "empty" class, denoted $c_{i}$. To enforce a target count $N_{target}$, we define a binary target vector $\mathbf{T} \in \mathbb{R}^{N_{max}}$ specifying which of the $N_{max}$ slots should be non-empty.

The guidance function is formulated as the Binary Cross-Entropy (BCEWithLogits) loss between the predicted "empty" logits and this target vector:
\begin{equation}
\label{equ:num_control}
\varphi_{\text{quantity}}(\mathbf{x}) = \text{BCEWithLogits}(\{c_{i}\}_{i=1}^{N_{max}}, \mathbf{T})
\end{equation}
The gradient of this function, $\nabla_{\mathbf{x}_t}\varphi_{\text{quantity}}$, provides a direct signal during sampling, steering the model to populate the scene with the $N_{target}$ desired objects.

\subsection{Articulated Collision Constraint}
\label{sec:control_articoll}

Standard collision losses are insufficient as they only check static geometry, ignoring functional plausibility. We introduce a differentiable guidance function, $\varphi_{\text{articoll}}$, to penalize such "functional collisions." Our method quantifies functional collision during diffusion sampling. For each object $b_i$ in the scene $\mathcal{B}$, we identify if it is articulated via a lookup. If so, we compute its \textit{functionally extended state}, $b'_i$, by heuristically expanding its bounding box along its primary axis of articulation. For non-articulated objects, $b'_i = b_i$. The total collision penalty is the sum of pairwise 3D Intersection over Union (IoU) between each object's extended state $b'_i$ and all other static objects $b_j$:
\begin{equation}
\label{equ:articoll}
\varphi_{\text{articoll}}(\mathbf{x}) = \sum_{i=1}^{N} \sum_{j=1, j \neq i}^{N} \text{IoU}_{3D}(b'_i, b_j)
\end{equation}
\indent This penalty is differentiable with respect to scene parameters. Its gradient, $\nabla_{\mathbf{x}_t}\varphi_{\text{articoll}}$, steers the reverse diffusion process away from obstructed configurations, ensuring generated scenes are functionally viable.

\subsection{Walkable Area Control}
\label{sec:control_walkable}

Ensuring a specific walkable space ratio is critical for robotic navigation. Enforcing such a constraint directly within the diffusion sampling loop would be computationally prohibitive, likely requiring expensive spatial queries at every step, and could potentially destabilize the generative process. We therefore introduce an efficient post-processing optimization as shown in Algorithm~\ref{alg:walkable_area_algo}, which decouples semantic layout generation from spatial density tuning, as shown in Figure~\ref{fig:model}. Our algorithm iteratively refines the scene to meet a target ratio $\tau$ by modifying only object \textit{sizes} while preserving their \textit{placements}. This strategy retains the core semantic structure while guaranteeing navigability.

\begin{algorithm}[t!]
\SetAlgoLined
\KwIn{Batch size $B$, max iterations $M$, ratio threshold $\tau$, top-$k$ objects $k$}
\For{$i \leftarrow 1$ \KwTo $B$}{
    Extract data for scene $i$\;
    $iter \leftarrow 0$\;
    $r_i \leftarrow$ CalculateWalkableRatio(scene $i$)\;

    \While{$r_i < \tau$ \textbf{and} $iter < M$}{
        $iter \leftarrow iter + 1$\;
        Sort valid objects by area\;

        $replacement \leftarrow$ False\;
        \For{top $k$ objects}{
            Find closest object in database\;
            \If{smaller replacement found}{
                Replace size and features\;
                $replacement \leftarrow$ True\;
            }
        }
        \If{not $replacement$}{
            \textbf{break}\;
        }
        $r_i \leftarrow$ CalculateWalkableRatio(scene $i$)\;
    }
}
\Return {optimized\_scenes}
\caption{Walkable Area Optimization} \label{alg:walkable_area_algo}
\end{algorithm}

\subsection{Proposed Task-Specific Evaluation Metrics}
\label{sec:evaluation metric 3}
To measure the controllability of Sec.~\ref{sec:llm_guide}, Sec.~\ref{sec:control_quantity}, Sec.~\ref{sec:control_articoll}, and Sec.~\ref{sec:control_walkable}, we proposed four novel evaluation metrics to validate them. Including LLM Controllability(Sec.~\ref{sec:llm controllability}), Object Quantity Controllability(Sec.~\ref{sec:object quantity controllability}), Articulated Object Collision Ratio(Sec.~\ref{sec:articulated object collision ratio}) and Walkable Area Controllability(Sec.~\ref{sec:walkable area controllability}). 
\subsubsection{LLM-Guided Layout Metric}
\label{sec:llm controllability}
To evaluate the structural and semantic fidelity of a generated graph $G_{\mathrm{gen}}=(V_{\mathrm{gen}}, E_{\mathrm{gen}})$ against the ground-truth $G_{\mathrm{gt}}=(V_{\mathrm{gt}}, E_{\mathrm{gt}})$, we measure node similarity, constraint satisfaction, and edge similarity.

\textbf{Node Similarity.} We compute a maximum cardinality matching $M: V_{\mathrm{gen}} \to V_{\mathrm{gt}}$ constrained by node type ($T(v_{\mathrm{gen}}) = T(v_{\mathrm{gt}})$). The score is the match size normalized by the larger graph size to penalize extraneous nodes:
\begin{equation}
S_{\mathrm{node}}(G_{\mathrm{gen}}, G_{\mathrm{gt}}) = \frac{|M|}{\max(|V_{\mathrm{gen}}|, |V_{\mathrm{gt}}|)}
\end{equation}

\textbf{Constraint Satisfaction Score.} This metric evaluates the area ratio distribution per room type ($R(G, c)$). We first measure the L1 distance between the generated and ground-truth distributions:
\begin{equation}
\label{eq:constraint_dist}
D_{L1} = \sum_{c \in \mathcal{C}} |R(G_{\mathrm{gen}}, c) - R(G_{\mathrm{gt}}, c)|
\end{equation}
The normalized constraint satisfaction score $S_{\mathrm{constraint}} \in $ is defined as:
\begin{equation}
\label{eq:constraint_score}
S_{\mathrm{constraint}}(G_{\mathrm{gen}}, G_{\mathrm{gt}}) = 1 - \frac{1}{2} D_{L1}
\end{equation}

\textbf{Edge Similarity.} Based on the node matching $M$, we identify the set of matched edges $E_{\mathrm{match}}$, where edges in $E_{\mathrm{gen}}$ have corresponding nodes (under $M$) that are also connected in $E_{\mathrm{gt}}$:
\begin{equation}
\label{eq:matched_edges}
E_{\mathrm{match}} = \{(u, v) \in E_{\mathrm{gen}} \mid (M(u), M(v)) \in E_{\mathrm{gt}}\}
\end{equation}
The score is normalized by the larger edge set to penalize spurious edges:
\begin{equation}
\label{eq:edge_score}
S_{\mathrm{edge}}(G_{\mathrm{gen}}, G_{\mathrm{gt}}) = \frac{|E_{\mathrm{match}}|}{\max(|E_{\mathrm{gen}}|, |E_{\mathrm{gt}}|)}
\end{equation}


\subsubsection{Object Quantity Control Metric}
\label{sec:object quantity controllability}
To assess the model's capability to control the number of objects within a generated scene, we conducted a quantitative evaluation. We prompted the model to generate rooms containing a specific target number of objects, $N_{\text{target}}$. A scene $S_i$ is considered a "success" if its generated object count, denoted $N(S_i)$, exactly matches the target. The Success Rate (SR) for a given target quantity over a set of $M$ scenes is then formally defined as:
\begin{equation}
\label{eq:quantity_success_rate}
SR(N_{\text{target}}) = \frac{1}{M} \sum_{i=1}^{M} \mathbb{I}(N(S_i) = N_{\text{target}})
\end{equation}
where $\mathbb{I}(\cdot)$ is the indicator function. For each target quantity, we calculated this rate over $M=100$ generated scenes. The results, presented in Table~\ref{tab:object_quantity}, show that our model maintains a high SR, consistently above 95\% for most tested quantities, demonstrating precise control over scene composition.

\subsubsection{Articulation Collision Metric}
\label{sec:articulated object collision ratio}
To quantitatively validate the effectiveness of our proposed articulated object collision constraint, $\varphi_{\text{articoll}}$, we conduct an evaluation by comparing our guided model against a baseline variant trained without this constraint. For each model, we generate a test set of 100 scenes and post-process them by transforming all articulated objects into their functionally extended states to simulate real-world usage. 

We introduce $R_{\text{acoll}}$ as our primary metric, defined as the proportion of articulated objects involved in functional collisions. Formally, for a given scene $S$, it is calculated as:
\begin{equation}
\label{eq:acoll_metric}
R_{\text{acoll}}(S) = \frac{1}{N_A} \sum_{j \in \mathcal{A}} \mathbb{I}\left( \max_{i \neq j} \left( \text{IoU}_{3D}(b'_i, b'_j) \right) > 0 \right)
\end{equation}
where $\mathcal{A}$ is the set of articulated objects in the scene ($N_A = |\mathcal{A}|$), and $b'_i, b'_j$ represent the functional bounding boxes. For an articulated object, this is its volume in the extended state; for a static object, it is its original bounding box. The indicator function $\mathbb{I}(\cdot)$ returns 1 if the condition is true. A lower $R_{\text{acoll}}$ score indicates superior functional plausibility.

\subsubsection{Walkable Area Controllability Metric}
\label{sec:walkable area controllability}
To quantitatively measure the navigability and spaciousness of a generated scene, we define the $R_{\text{walkable}}$. This metric is calculated as the ratio of the total unobstructed floor area to the total area of the room. Let $A_{\text{room}}$ be the total area of the floor plan. For a scene containing $N$ objects, where the floor footprint of the $i$-th object is denoted by $A_i$, the total walkable area $A_{\text{walkable}}$ is the room area minus the sum of all object footprints. The ratio is formally defined as:

\begin{equation}
\label{eq:walkable_ratio}
R_{\text{walkable}} = \frac{A_{\text{walkable}}}{A_{\text{room}}} = \frac{A_{\text{room}} - \sum_{i=1}^{N} A_i}{A_{\text{room}}}
\end{equation}

The SR over a set of $M$ generated scenes for a given threshold is then formally defined as:
\begin{equation}
\label{eq:success_rate}
SR(\tau_{\text{walkable}}) = \frac{1}{M} \sum_{i=1}^{M} \mathbb{I}(R_{\text{walkable}}(S_i) \ge \tau_{\text{walkable}})
\end{equation}
\section{Experiment}
\label{sec:experiment}

\subsection{Implementation Detail}
\textbf{Datasets.}
Our pipeline uses three specialized datasets. The layout generator is trained on 3D-FRONT \cite{fu20213dfront3dfurnishedrooms} with 14,629 indoor scenes, and layouts are populated with textured assets from 3D-FUTURE \cite{fu20203dfuture3dfurnitureshape} containing 16,563 furniture models. To enable interactivity, we use GAPartNet \cite{geng2023gapartnetcrosscategorydomaingeneralizableobject}, which provides part-level semantics and pose data for 8,489 parts across 1,166 objects.

\textbf{Baselines.}
We compare with three baselines to demonstrate methodological progression. ATISS \cite{paschalidou2021atissautoregressivetransformersindoor} is an autoregressive Transformer for sequential object generation, DiffuScene \cite{tang2024diffuscenedenoisingdiffusionmodels} employs diffusion to improve global consistency, and PhyScene \cite{yang2024physcenephysicallyinteractable3d} adds physics-based guidance for physically plausible scene synthesis.

\textbf{Evaluation Metrics.}
We assess layout quality using Fréchet Inception Distance (FID)  \cite{heusel2018ganstrainedtimescaleupdate}, Kernel Inception Distance (KID) \cite{bińkowski2021demystifyingmmdgans}, Scene Classification Accuracy (SCA), and Category KL divergence (CKL) following \cite{tang2024diffuscenedenoisingdiffusionmodels}. Constraint-specific metrics are defined in Sec.~\ref{sec:evaluation metric 3} and evaluated in Sec.~\ref{sec:LLM Controllability 4}–\ref{sec:Walkable Area Controllability 4}.

\begin{figure*}[t!]
    \centering
    \includegraphics[width=1.0\linewidth]{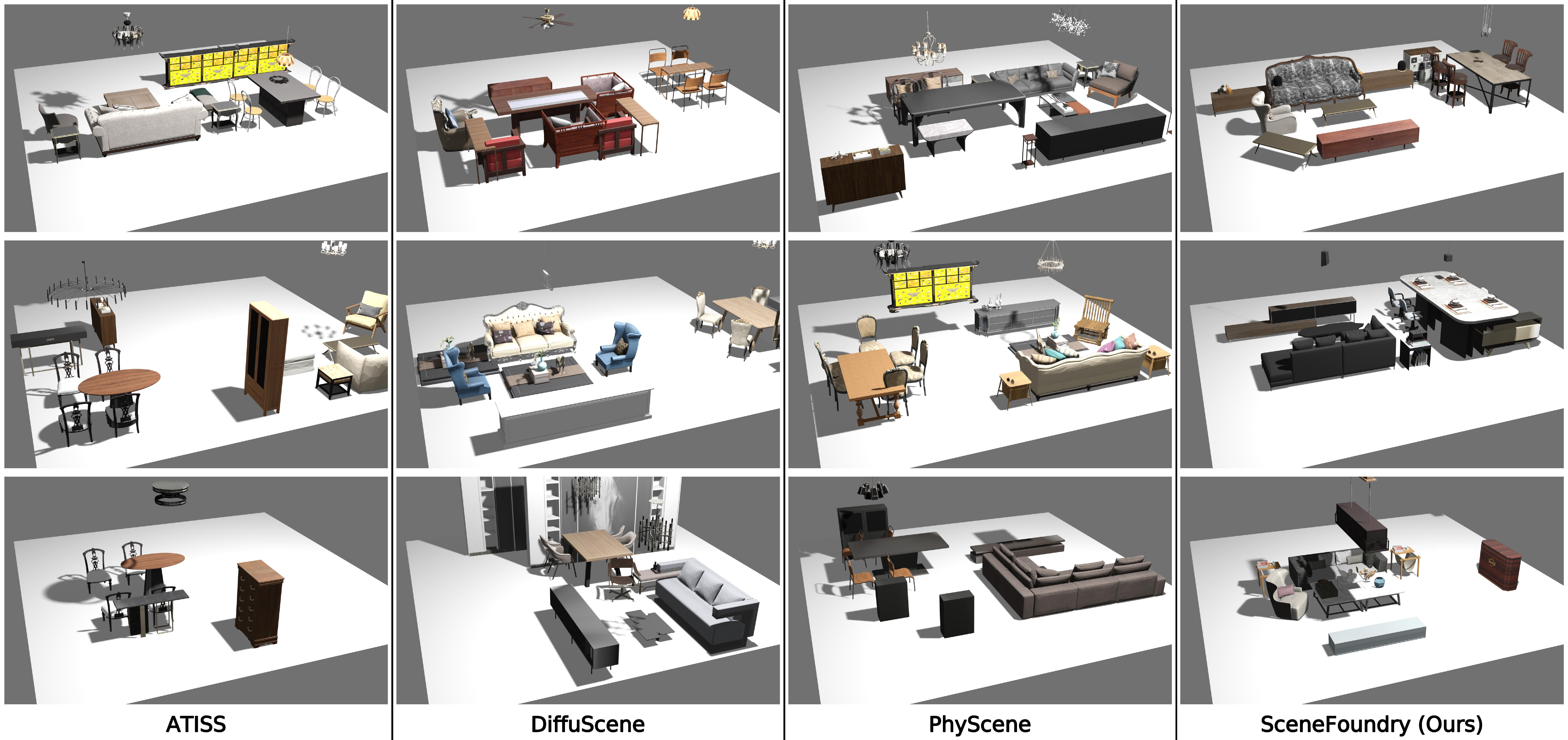}
    \caption{Qualitative comparison of conditioned scene synthesis results among PhyScene, ATISS, DiffuScene, and SceneFoundry.}
    \label{fig:visual_condition_result}
\end{figure*}

\subsection{Conditioned Scene Synthesis Evaluation} 
\label{sec:condition_scene}
Conditioned scene synthesis is evaluated with diverse textual and spatial prompts. As shown in Figure~\ref{fig:visual_condition_result}, SceneFoundry generates layouts that align with user-defined conditions and maintain spatial coherence. Quantitative results in Table~\ref{tab:generation_results} show high structural realism, minimal artifacts, and superior KID and CKL scores, confirming controllable high-fidelity scene generation.

\begin{table}[htbp]
\centering
\caption{\textbf{Floor-conditioned Scene Synthesis.} We compare SceneFoundry with baseline on common perceptual quality scores FID, KID, SCA, CKL.}
\vspace{-5pt}
\resizebox{\columnwidth}{!}{
    \begin{tabular}{ccccc}
        \toprule
        Method & FID $\downarrow$ & KID $\downarrow$ & SCA & CKL $\downarrow$ \\ 
        \midrule
          
        ATISS  &  30.19  &  0.0010  & 49.14& 0.0028 \\
        DiffuScene  & \textbf{25.00} & \textbf{0.0004}  & 51.78 &0.0031 \\
        PhyScene & 25.52 & 0.0006 & \textbf{50.10} & 0.0025 \\
        \midrule
        SceneFoundry & 29.02 & \textbf{0.0004} & 49.11 & \textbf{0.0024} \\
        \bottomrule
    \end{tabular}
}
\label{tab:generation_results}
\end{table}

\subsection{LLM-Guided Layout Generation Evaluation}
\label{sec:LLM Controllability 4}
The resulting layouts were quantitatively evaluated against ground-truth graphs that perfectly satisfy the given high-level constraints. We use our proposed room graph similarity metrics ($S_{\text{node}}$, $S_{\text{constraint}}$, and $S_{\text{edge}}$) from Sec.~\ref{sec:llm controllability} to measure the semantic and structural fidelity of the generated floor plans. Our method achieves extremely high scores across all three metrics as shown in Table~\ref{tab:llm_results}.

\begin{table}[htbp]
\caption{Results for our LLM Controllability experiment.}
\centering
\begin{tabular}{lccc}
\toprule
\textbf{Method} &$S_{\text{node}}$ $\uparrow$ &$S_{\text{constraint}}$ $\uparrow$ &$S_{\text{edge}}$ $\uparrow$ \\
\midrule
Ours (LLM control) & \textbf{0.989} & \textbf{0.923} & \textbf{0.954} \\
\bottomrule
\end{tabular}
\label{tab:llm_results}
\end{table}

\subsection{Object Quantity Controllability Evaluation}
\label{sec:Object Quantity Controllability 4}

Rooms are generated with target object counts $N_{\text{target}}$ ranging from 5 to 16. A generation is successful when the final count of non-empty slots matches $N_{\text{target}}$. For each target, 100 scenes are sampled, and the success rate (SR) is reported in Table~\ref{tab:object_quantity}. Results show consistently high SR values (0.95–0.97), demonstrating stable quantity control and robustness to scene complexity.

\begin{table}[htbp]
\caption{SR of generating scenes with a specific target number of objects. The rate is calculated over 100 generated scenes for each target.}
\centering
\begin{tabular}{cc @{\hspace{2em}} cc @{\hspace{2em}} cc}
\toprule
\textbf{$N_{target}$} & \textbf{SR} & \textbf{$N_{target}$} & \textbf{SR} & \textbf{$N_{target}$} & \textbf{SR} \\
\midrule
5 & 0.95 & 9 & 0.96 & 13 & 0.96 \\
6 & 0.95 & 10 & 0.97 & 14 & 0.95 \\
7 & 0.96 & 11 & 0.96 & 15 & 0.95 \\
8 & 0.95 & 12 & 0.96 & 16 & 0.95 \\
\bottomrule
\end{tabular}
\label{tab:object_quantity}
\end{table}

\begin{table}[htbp]
\caption{Comparison of $R_{\text{acoll}}$ and $R_{\text{reach}}$. Our method drastically reduces functional collisions ($R_{\text{acoll}} \downarrow$) and improves object accessibility ($R_{\text{reach}} \uparrow$).}
\centering
\setlength{\tabcolsep}{12pt} 
\begin{tabular*}{\columnwidth}{@{\extracolsep{\fill}}lcc@{}}
\toprule
\textbf{Method} & $R_{\text{acoll}} \downarrow$ & $R_{\text{reach}} \uparrow$ \\
\midrule
Baseline (w/o $\varphi_{\text{articoll}}$) & 0.191 & 0.742 \\
Ours (w/ $\varphi_{\text{articoll}}$) & \textbf{0.109} & \textbf{0.808} \\
\bottomrule
\end{tabular*}
\label{tab:articoll_results}
\end{table}

\begin{figure*}[t!]
    \centering
    \includegraphics[width=1.0\linewidth]{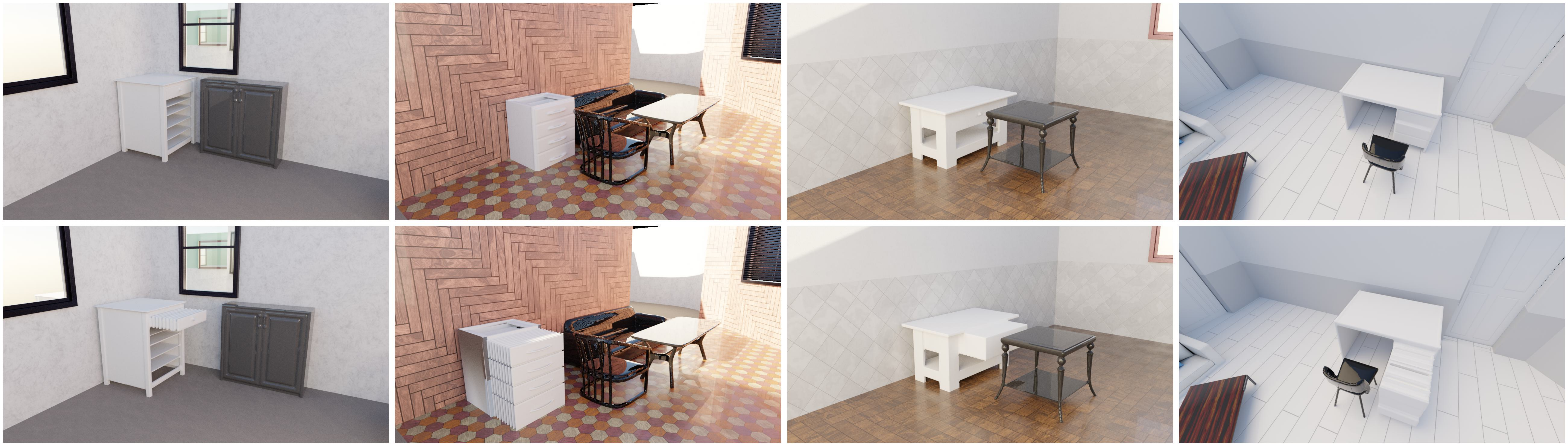}
    \caption{Visualization of the Articulated Object Collision Constraint. Synthesized scenes without the constraint (top) show obstructed articulated furniture, such as drawers that cannot open, while applying the constraint (bottom) enables proper motion and functional layouts.}
    \label{fig:ex_walkable_area_and_articollision}
\end{figure*}

\subsection{Articulated Collision Constraint Evaluation}
\label{sec:Object Articulated Controllability 4}
We further evaluate the effect of the proposed Articulated Object Collision Constraint, which enforces functional clearance for movable furniture. As illustrated in Figure~\ref{fig:ex_walkable_area_and_articollision}, scenes generated without this constraint often contain obstructed articulated parts, such as drawers or chairs that cannot move freely. When the constraint is applied, these collisions are effectively eliminated, resulting in functionally usable layouts. Quantitatively, our method achieves a significantly lower functional collision rate $R_{\text{acoll}}$ (Sec.~\ref{sec:articulated object collision ratio}) and higher object reachability ($R_{\text{reach}}$) \cite{yang2024physcenephysicallyinteractable3d} than the baseline, as shown in Table~\ref{tab:articoll_results}. The constraint improves scene functionality and accessibility in generated scenes.


\subsection{Walkable Area Controllability Evaluation}
\label{sec:Walkable Area Controllability 4}
\begin{figure}[htbp]
    \centering
    \includegraphics[width=1.0\linewidth]{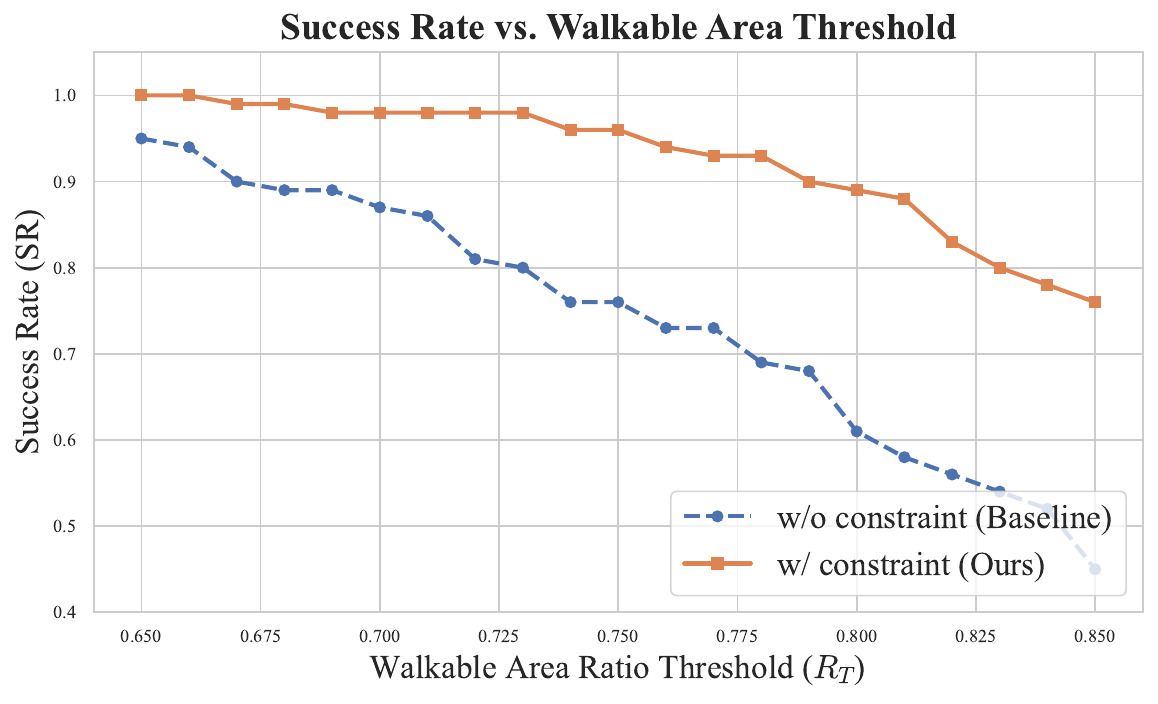}
    \caption{
        Success Rate (SR) versus Walkable Area Ratio Threshold ($R_T$). Walkable Area Control (orange) consistently outperforms the baseline (blue), ensuring navigable layouts.
    }
    \label{fig:walkable_area_comparison}
\end{figure}

\begin{figure}[htbp]
    \centering
    \includegraphics[width=1.0\linewidth]{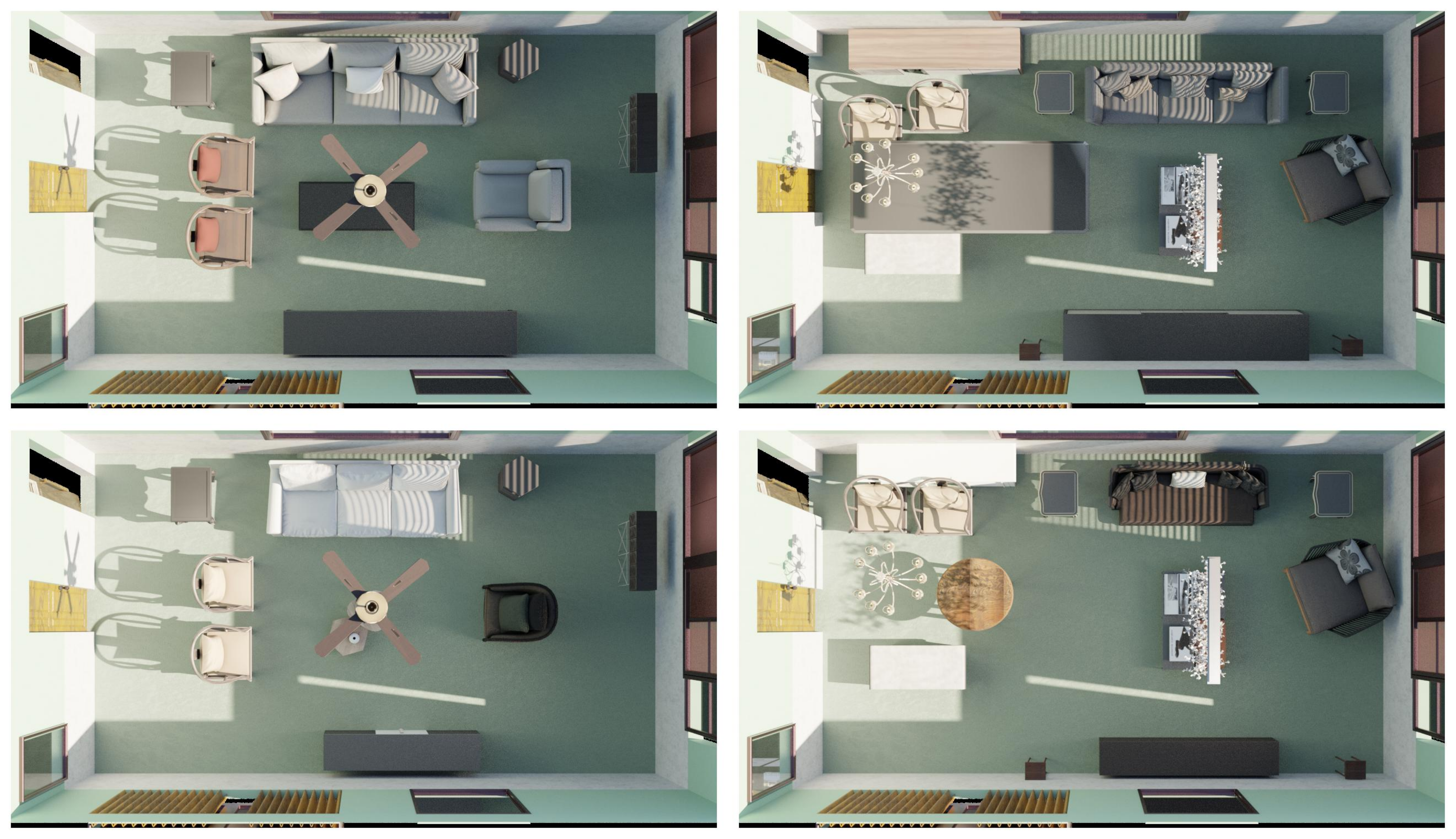}
    \caption{Visualization of Walkable Area Control.}
    \label{fig:walkable_result}
\end{figure}

Thresholds from 0.60 to 0.95 were tested with $M=100$ scenes. For each threshold, we compare the SR under two conditions: a baseline without our constraint and with our constraint activated. Our method significantly increases the SR across all tested thresholds, as shown in Figure~\ref{fig:walkable_area_comparison}. Qualitative examples in Figure~\ref{fig:walkable_result} further show that the constraint maintains sufficient free space for navigation while preserving realistic scene density.

\subsection{Ablation Study on Scene Generation Control}
We conduct an ablation study to validate the effectiveness of our proposed guidance mechanisms for controlling scene plausibility. Following the evaluation protocol established by PhyScene \cite{yang2024physcenephysicallyinteractable3d}, we measure three key metrics: object collision ($\mathbf{Col}_{\text{obj}}\downarrow$), walkable ratio ($\mathbf{R_{\text{walkable}}}\uparrow$), and object reachability ($\mathbf{R}_{\text{reach}}\uparrow$). We analyze the contributions of our two guidance functions: ArtiCollision and Walkable Ratio, as shown in Table~\ref{tab:ablation study}.

\begin{table}[htbp]
\caption{Ablation study on the use of guidance functions. Our final result balances the effectiveness of the two guidance mechanisms.}
\vspace{-5pt}
\resizebox{\linewidth}{!}{
\begin{tabular}{cccccc}
\toprule
ArtiCollision & Walkable Ratio  &  $\mathbf{Col}_{\text{obj}}\downarrow$ & $\mathbf{R_{\text{walkable}}}\uparrow$     & $\mathbf{R}_{\text{reach}}\uparrow$    \\ \midrule
            &            &  0.279  & 0.774  & 0.742   \\
 \checkmark &            &  0.267  & 0.774  & 0.808  \\
            & \checkmark &  0.250  & 0.822  & 0.782  \\
 \midrule
 \checkmark & \checkmark &  \textbf{0.249}  & \textbf{0.822}  & \textbf{0.830}  \\
 \bottomrule
\end{tabular}
}
\label{tab:ablation study}
\end{table}

\section{Conclusion}
This paper presents a multi-stage framework for controllable 3D environment generation that connects user intent with structured scene synthesis. The method integrates an LLM-guided floor-plan generator and a diffusion-based layout model to achieve semantic and spatial coherence. Functional modules including Object Quantity Control, Articulated Object Collision, and Walkable Area Control ensure realistic, accessible, and navigable layouts. Experimental results demonstrate precise control, high perceptual quality, and strong functional plausibility, consistently outperforming baseline methods and establishing a solid foundation for future Sim-to-Real research.

\cleardoublepage

{
    \small
    \bibliographystyle{ieeenat_fullname}
    \bibliography{main}
}

\newgeometry{
   left=3cm,
   right=3cm
}

\twocolumn[
  \begin{center}
    {\LARGE \textbf{Appendix}} \\ 
    \vspace{15pt}
    {\large \textbf{\textsc{SceneFoundry}: Generating Interactive Infinite 3D Worlds}}
    \vspace{10pt}
  \end{center}
]
\setcounter{section}{0}
\renewcommand{\thesection}{\Alph{section}}
\vspace{1cm}
\section{Reproducibility and Code Release}
To ensure the reproducibility of our results and facilitate future research in controllable scene generation, we will release our complete source code, trained model weights, and the post-processing scripts upon acceptance. Detailed instructions for environment setup and inference are provided in the supplementary material.

\noindent
\small
\url{https://github.com/anc891203/SceneFoundry}

\section{Discussion}
We provide a critical analysis of the current limitations of our \textbf{SceneFoundry} framework and discuss the broader societal implications of our work below.

\subsection{Limitation}
Despite the demonstrated efficacy of SceneFoundry in generating functionally viable and apartment-scale environments, several limitations remain to be addressed. 

\textbf{Inference Latency.} A primary constraint is the computational cost associated with the multi-stage pipeline. While the LLM-guided floor plan generation is relatively efficient, the core diffusion-based furniture population relies on iterative denoising steps. Coupled with the gradient calculations required for our novel constraints, the inference time for a full apartment scale is considerable. This currently precludes the system from real-time generation applications.

\textbf{Heuristic Approximation of Articulation.} Our Articulated Object Collision Constraint relies on a heuristic expansion of bounding boxes to approximate the kinematic workspace of objects. While robust for standard furniture morphologies found in GAPartNet, this axis-aligned expansion simplifies the complex, potentially non-linear trajectories of certain articulated parts. Consequently, for highly complex mechanisms or multi-jointed objects, the collision avoidance might be overly conservative or, in rare cases, insufficient.

\textbf{Dataset Bias and Generalization.} The stylistic and semantic diversity of our generated scenes is inherently bounded by the underlying training data, specifically 3D-FRONT and GAPartNet. While these datasets are extensive, they may not fully encompass the architectural styles of different cultures or historical periods. As with all learning-based generative models, the system may exhibit biases present in the dataset, potentially favoring modern, Western-style interior layouts over others.

\subsection{Social Impact}
The primary societal contribution of this work lies in its potential to accelerate the development of embodied AI and service robotics. By automating the synthesis of large-scale, functionally sound training environments, SceneFoundry significantly reduces the reliance on costly and labor-intensive real-world data collection. This democratization of high-quality simulation data can foster innovation in domestic robotics, potentially leading to deploying intelligent agents capable of assisting the elderly or individuals with disabilities in their daily lives.

From an environmental perspective, while the training of large diffusion models incurs a carbon footprint, the ability to train robots in simulation (Sim-to-Real) drastically reduces the energy consumption, material waste, and physical risks associated with trial-and-error learning in the physical world. Regarding ethical considerations, unlike generative models for faces or media, the generation of indoor scene layouts carries a relatively low risk of malicious misuse. However, we advocate for continued awareness regarding the cultural biases embedded in synthetic datasets to ensure the inclusivity of future AI technologies.

\section{Preliminaries}
Our generative framework is built upon Denoising Diffusion Probabilistic Models (DDPMs) \cite{ho2020denoisingdiffusionprobabilisticmodels}. DDPMs are a class of latent variable models designed to learn a data distribution $p(\mathbf{x})$ by reversing a gradual noising process. In this section, we briefly review the mathematical formulation of DDPMs, including the forward diffusion process, the reverse denoising process, and the training objective.

\subsection{Denoising Diffusion Probabilistic Models}
\label{sec:ddpm_background}

\textbf{Forward Diffusion Process.}
The forward process, also known as the diffusion process, is a fixed Markov chain that gradually adds Gaussian noise to the data $\mathbf{x}_0 \sim q(\mathbf{x}_0)$ over a sequence of timesteps $t = 1, \dots, T$. The transition probability at each step is defined as:
\begin{equation}
    q(\mathbf{x}_t | \mathbf{x}_{t-1}) = \mathcal{N}(\mathbf{x}_t; \sqrt{1 - \beta_t} \mathbf{x}_{t-1}, \beta_t \mathbf{I}),
\end{equation}
where $\{\beta_t \in (0, 1)\}_{t=1}^T$ is a pre-defined variance schedule. As $T \to \infty$, the data $\mathbf{x}_T$ approaches an isotropic Gaussian distribution $\mathcal{N}(\mathbf{0}, \mathbf{I})$. A key property of this process is that we can sample $\mathbf{x}_t$ at any arbitrary timestep $t$ directly from $\mathbf{x}_0$ in closed form:
\begin{equation}
    q(\mathbf{x}_t | \mathbf{x}_0) = \mathcal{N}(\mathbf{x}_t; \sqrt{\bar{\alpha}_t} \mathbf{x}_0, (1 - \bar{\alpha}_t) \mathbf{I}),
\end{equation}
where $\alpha_t = 1 - \beta_t$ and $\bar{\alpha}_t = \prod_{s=1}^t \alpha_s$. This allows us to express $\mathbf{x}_t$ as a linear combination of the original data and noise:
\begin{equation}
    \mathbf{x}_t = \sqrt{\bar{\alpha}_t}\mathbf{x}_0 + \sqrt{1 - \bar{\alpha}_t}\bm{\epsilon}, \quad \bm{\epsilon} \sim \mathcal{N}(\mathbf{0}, \mathbf{I}).
\end{equation}

\textbf{Reverse Denoising Process.}
The goal of the generative model is to reverse this process, sampling from $q(\mathbf{x}_{t-1}|\mathbf{x}_t)$ to reconstruct the data. Since the exact posterior is intractable, we approximate it using a learnable Markov chain with parameterized Gaussian transitions:
\begin{equation}
    p_\theta(\mathbf{x}_{t-1} | \mathbf{x}_t) = \mathcal{N}(\mathbf{x}_{t-1}; \bm{\mu}_\theta(\mathbf{x}_t, t), \bm{\Sigma}_\theta(\mathbf{x}_t, t)),
\end{equation}
starting from $\mathbf{x}_T \sim \mathcal{N}(\mathbf{0}, \mathbf{I})$. The mean $\bm{\mu}_\theta$ and covariance $\bm{\Sigma}_\theta$ are predicted by neural networks. Following \cite{ho2020denoisingdiffusionprobabilisticmodels}, we set $\bm{\Sigma}_\theta(\mathbf{x}_t, t) = \sigma_t^2 \mathbf{I}$, where $\sigma_t^2$ is set to $\beta_t$ or $\tilde{\beta}_t = \frac{1-\bar{\alpha}_{t-1}}{1-\bar{\alpha}_t}\beta_t$. The mean is parameterized to predict the noise $\bm{\epsilon}$ added to $\mathbf{x}_0$:
\begin{equation}
    \bm{\mu}_\theta(\mathbf{x}_t, t) = \frac{1}{\sqrt{\alpha_t}} \left( \mathbf{x}_t - \frac{\beta_t}{\sqrt{1 - \bar{\alpha}_t}} \bm{\epsilon}_\theta(\mathbf{x}_t, t) \right).
\end{equation}

\textbf{Optimization Objective.}
The model is trained by optimizing the variational lower bound on the negative log-likelihood. \citet{ho2020denoisingdiffusionprobabilisticmodels} demonstrated that a simplified objective yields better sample quality. This simplified loss calculates the mean squared error between the true noise $\bm{\epsilon}$ and the predicted noise $\bm{\epsilon}_\theta$:
\begin{equation}
    \mathcal{L}_{\text{simple}} = \mathbb{E}_{t, \mathbf{x}_0, \bm{\epsilon}} \left[ \| \bm{\epsilon} - \bm{\epsilon}_\theta(\mathbf{x}_t, t) \|^2 \right],
\end{equation}
where $t$ is uniformly sampled from $\{1, \dots, T\}$. In our framework, we adapt this backbone to generate 3D scene layouts by conditioning the denoiser on floor plan constraints.

\section{Implementation Details}
\textbf{Experimental Setting.} We evaluate our method on the 3D-FRONT dataset, employing the official train/test splits to ensure consistency with prior work. For articulation-aware generation, we augment the object assets using the GAPartNet dataset, which provides part-level annotations. To verify the robustness of our method, all baselines are retrained on this identical data subset. We generate 1,000 scenes for each experimental condition to compute reliable metrics.

\textbf{Evaluation Metrics.} To quantitatively evaluate the quality, diversity, and semantic coherence of our generated scenes, we employ a comprehensive suite of metrics. 
\textit{Standard Perceptual \& Semantic Metrics:}
\begin{itemize}
    \item \textbf{Fr\'echet Inception Distance (FID)~\cite{heusel2018ganstrainedtimescaleupdate}:} Measures the distributional distance between deep features of generated ($\mu_g, \Sigma_g$) and real ($\mu_r, \Sigma_r$) scene renderings:
    \begin{equation}
        \text{FID} = \|\mu_r - \mu_g\|^2 + \text{Tr}(\Sigma_r + \Sigma_g - 2(\Sigma_r \Sigma_g)^{1/2})
    \end{equation}
    
    \item \textbf{Kernel Inception Distance (KID)~\cite{bińkowski2021demystifyingmmdgans}:} An unbiased estimator of the Maximum Mean Discrepancy (MMD) between feature representations, suitable for smaller sample sizes:
    \begin{equation}
    \begin{split}
        \text{KID} = &\text{MMD}^2(P_r, P_g) \\
        & = \mathbb{E}[k(x, x')] + \mathbb{E}[k(y, y')] - 2\mathbb{E}[k(x, y)]
    \end{split}
    \end{equation}
    
    \item \textbf{Scene-Class Alignment (SCA)~\cite{tang2024diffuscenedenoisingdiffusionmodels}:} Evaluates semantic consistency by calculating the classification accuracy of a pre-trained scene classifier $C$ on generated layouts $x_{gen}$:
    \begin{equation}
        \text{SCA} = \mathbb{E}_{x_{gen}} [\mathbb{I}(C(x_{gen}) = y_{label})]
    \end{equation}
    
    \item \textbf{Category KL divergence (CKL)~\cite{tang2024diffuscenedenoisingdiffusionmodels}:} Measures the divergence between the object category distribution of the generated set ($P_g$) and the ground truth ($P_r$):
    \begin{equation}
        \text{CKL} = D_{KL}(P_g || P_r) = \sum_{i} P_g(i) \log \frac{P_g(i)}{P_r(i)}
    \end{equation}
\end{itemize}
\textit{Proposed Controllability Metrics (Ours):}
\begin{itemize}
    \item \textbf{LLM-Guided Layout Metric:} Evaluates the structural and semantic fidelity of the generated floor plan graph against the ground truth by assessing node matching, edge connectivity, and constraint satisfaction.
    \item \textbf{Object Quantity Control Metric:} Defines the success rate of generating scenes that contain the exact target number of objects specified by the user.
    \item \textbf{Articulation Collision Ratio:} Measures the percentage of articulated objects (e.g., cabinets) that are functionally obstructed by other objects when in their open/extended state.
    \item \textbf{Walkable Area Controllability:} Calculates the success rate of generating scenes where the ratio of unobstructed walkable floor area meets or exceeds a specified threshold.
\end{itemize}

\subsection{Compare Model Settings}
We benchmark SceneFoundry against three state-of-the-art baselines:
\begin{itemize}
    \item \textbf{ATISS:} An autoregressive transformer model that places objects sequentially. We use the official implementation, retraining it on our dataset split for fair comparison.
    \item \textbf{DiffuScene:} A diffusion-based model that generates scene layouts in parallel. This represents the current state-of-the-art in unconditional layout generation.
    \item \textbf{PhyScene:} A recent physics-aware generative model. We compare against PhyScene to highlight the advantages of our specific articulated object constraints.
\end{itemize}
All models receive the same floor plan input during the conditional generation tasks.

\subsection{Training Details}
The model is trained using the Adam optimizer with a learning rate of $2 \times 10^{-4}$ and weight decay of 0.0. We employ a step learning rate schedule with a step size of 20,000 and a decay factor of 0.5. Training runs for 130,000 epochs with a batch size of 128. The gradient norm is clipped at 10. Table~\ref{tab:training_hyperparams} summarizes the complete hyperparameter configuration.

\begin{table}[h]
\centering
\caption{Detailed hyperparameter settings for training the diffusion model.}
\label{tab:training_hyperparams}
\begin{tabular}{lc}
\toprule
\textbf{Configuration} & \textbf{Value} \\
\midrule
Optimizer & Adam \\
Base Learning Rate & $2 \times 10^{-4}$ \\
Weight Decay & 0.0 \\
Batch Size & 128 \\
Max Gradient Norm & 10 \\
\midrule
Learning Rate Schedule & Step Decay \\
LR Step Size & 20,000 \\
LR Decay Factor ($\gamma$) & 0.5 \\
Total Epochs & 130,000 \\
\bottomrule
\end{tabular}
\end{table}

\subsection{Computing Resource Configuration}
All model training and evaluation were conducted on a computing node equipped with a single \textbf{NVIDIA 3090 GPU (24GB VRAM)} and an \textbf{Intel Core i9-12900K}. Under this configuration, training the core diffusion model takes approximately 1500 hours. During inference, generating a complete apartment-scale scene (3 rooms) with full constraint guidance takes approximately 300 seconds per scene.

\section{Additional Experiments}
\subsection{LLM Controllability}
To rigorously evaluate the fidelity of our LLM-based parameter space guidance, we designed a comprehensive benchmark consisting of 20 distinct natural language prompts.

\textbf{Overall Performance.} 
We define a generation as successful only if the final layout strictly adheres to all constraints specified in the prompt, as shown in Table~\ref{tab:llm_summary}.

\begin{table}[htbp]
\centering
\caption{Summary of LLM Controllability Experiments.}
\label{tab:llm_summary}
\begin{tabular}{lr}
\toprule
\textbf{Metric} & \textbf{Value} \\
\midrule
Total Test Cases & 20 \\
Average Score & 96.5\% \\
\bottomrule
\end{tabular}
\end{table}

\textbf{Component Analysis.} 
To understand the specific challenges in layout control, we decompose the performance into three sub-metrics. The score is composed of weights in the Table~\ref{tab:llm_component}.
\begin{itemize}
    \item \textbf{Room Presence:} Existence of required room types.
    \item \textbf{Adjacency:} Correct connectivity between rooms.
    \item \textbf{Constraints:} Geometric or functional requirements.
\end{itemize}

\begin{table}[htbp]
\centering
\caption{Weights indicate the relative importance assigned to each component in the overall score.}
\label{tab:llm_component}
\begin{tabular}{lrr}
\toprule
\textbf{Component} & \textbf{Avg. Score} & \textbf{Weight} \\
\midrule
Room Presence & 98.9\% & 50\% \\
Adjacency & 92.3\% & 30\% \\
Constraints & 95.4\% & 20\% \\
\bottomrule
\end{tabular}
\end{table}

\textbf{Detailed Test Results}
We provide a granular breakdown of each test case in Table~\ref{tab:long_results}. In these cases, the system maintains high scores on connectivity and functional constraints, ensuring the generated layouts remain usable.

\begin{table*}[t]
\centering
\caption{Complete Test Results for LLM Controllability. This table details the performance of 20 distinct test prompts.}
\label{tab:long_results}
\begin{tabular}{lrrrrr}
\toprule
\textbf{Test ID} & \textbf{Score} & \textbf{Time(s)}& \textbf{Room Presence} & \textbf{Adjacency} & \textbf{Constraints} \\
\midrule
two\_bedroom\_apt\_01 & 100.0\% & 262.2 & 100\% & 100\% & 100\% \\
open\_plan\_loft\_01 & 100.0\% & 274.7 & 100\% & 100\% & 100\% \\
family\_home\_01 & 100.0\% & 276.7 & 100\% & 100\% & 100\% \\
master\_suite\_home\_01 & 100.0\% & 265.3 & 100\% & 100\% & 100\% \\
four\_bedroom\_house\_01 & 100.0\% & 270.8 & 100\% & 100\% & 100\% \\
guest\_suite\_home\_01 & 100.0\% & 282.2 & 100\% & 100\% & 100\% \\
dual\_master\_suite\_01 & 100.0\% & 269.5 & 100\% & 100\% & 100\% \\
balcony\_apartment\_01 & 100.0\% & 285.2 & 100\% & 100\% & 100\% \\
multigenerational\_home\_01 & 100.0\% & 289.7 & 100\% & 100\% & 100\% \\
basic\_studio\_01 & 100.0\% & 259.0 & 100\% & 100\% & 100\% \\
one\_bedroom\_apt\_01 & 100.0\% & 270.0 & 100\% & 100\% & 100\% \\
compact\_efficiency\_01 & 100.0\% & 269.7 & 100\% & 100\% & 100\% \\
student\_apartment\_01 & 100.0\% & 266.0& 100\% & 100\% & 100\% \\
single\_floor\_accessible\_01 & 100.0\% & 262.5 & 100\% & 100\% & 100\% \\
entertainment\_home\_01 & 95.8\% & 274.9 & 100\% & 100\% & 79.2\% \\
work\_from\_home\_01 & 90.0\% & 263.0 & 80.0\% & 100\% & 100\% \\
luxury\_penthouse\_01 & 88.8\% & 270.8 & 87.5\% & 100\% & 75.0\% \\
three\_bed\_townhouse\_01 & 85.0\% & 268.3& 100\% & 50\% & 100\% \\
separated\_zones\_01 & 85.0\% & 285.7 & 100\% & 50.0\% & 100\% \\
home\_office\_layout\_01 & 62.5\% & 267.8 & 25.0\% & 100\% & 100\% \\
\bottomrule
\end{tabular}
\end{table*}

\section{Render Results}

We present an extensive qualitative evaluation of the 3D indoor layouts generated by our proposed method. While quantitative metrics provide numerical evidence of our model's performance, visual inspection is equally crucial for assessing the perceptual quality, spatial coherence, and practical usability of the synthesized scenes. To this end, we provide a comprehensive gallery of results across a diverse range of scene categories, demonstrating the robustness of our approach in handling complex room geometries.

To guarantee the highest visual fidelity, all visualizations were produced using the \textbf{Blender} creation suite, leveraging its advanced \textbf{Cycles} rendering engine. Cycles is a production-grade, physically-based path tracer that excels at simulating the intricate interactions of light transport. Unlike real-time rasterization engines, Cycles calculates global illumination, multi-bounce indirect lighting, and accurate soft shadows, which are essential for verifying that objects are properly grounded and not floating. Furthermore, we utilized high-resolution Physically Based Rendering textures and materials to enhance the realism of the furniture, allowing for a rigorous assessment of the layout's aesthetic quality.

The rendering pipeline imposes significant computational demands, particularly when processing scenes with high-poly assets and complex lighting setups. Consequently, all rendering tasks were executed on a dedicated workstation equipped with an \textbf{Intel Core i3-14100F CPU} and an \textbf{NVIDIA RTX 5090 GPU}. The massive 32GB VRAM of the RTX 5090 proved instrumental in loading large-scale scene data and high-resolution textures without memory bottlenecks, while the CPU efficiently managed scene graph traversal and asset loading.

\begin{figure*}[t]
    \centering
    \includegraphics[width=1.0\linewidth]{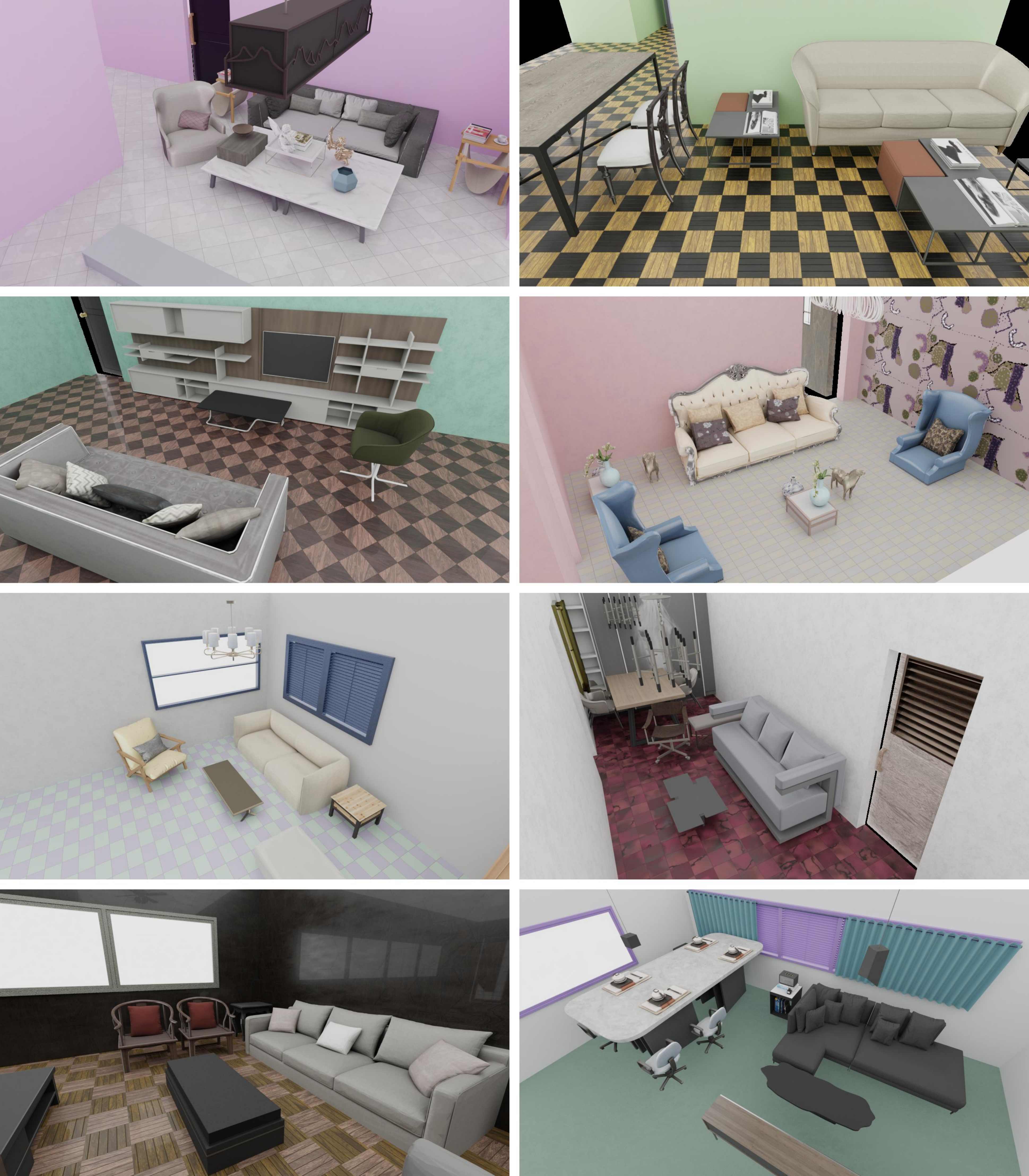}
    \caption{\textbf{Generated 3D Layouts.} Representative visualization of scenes generated by SceneFoundry. (Part 1).}
    \label{fig:render_result_1}
    \vspace{1cm}
\end{figure*}

\begin{figure*}[t]
    \centering
    \includegraphics[width=1.0\linewidth]{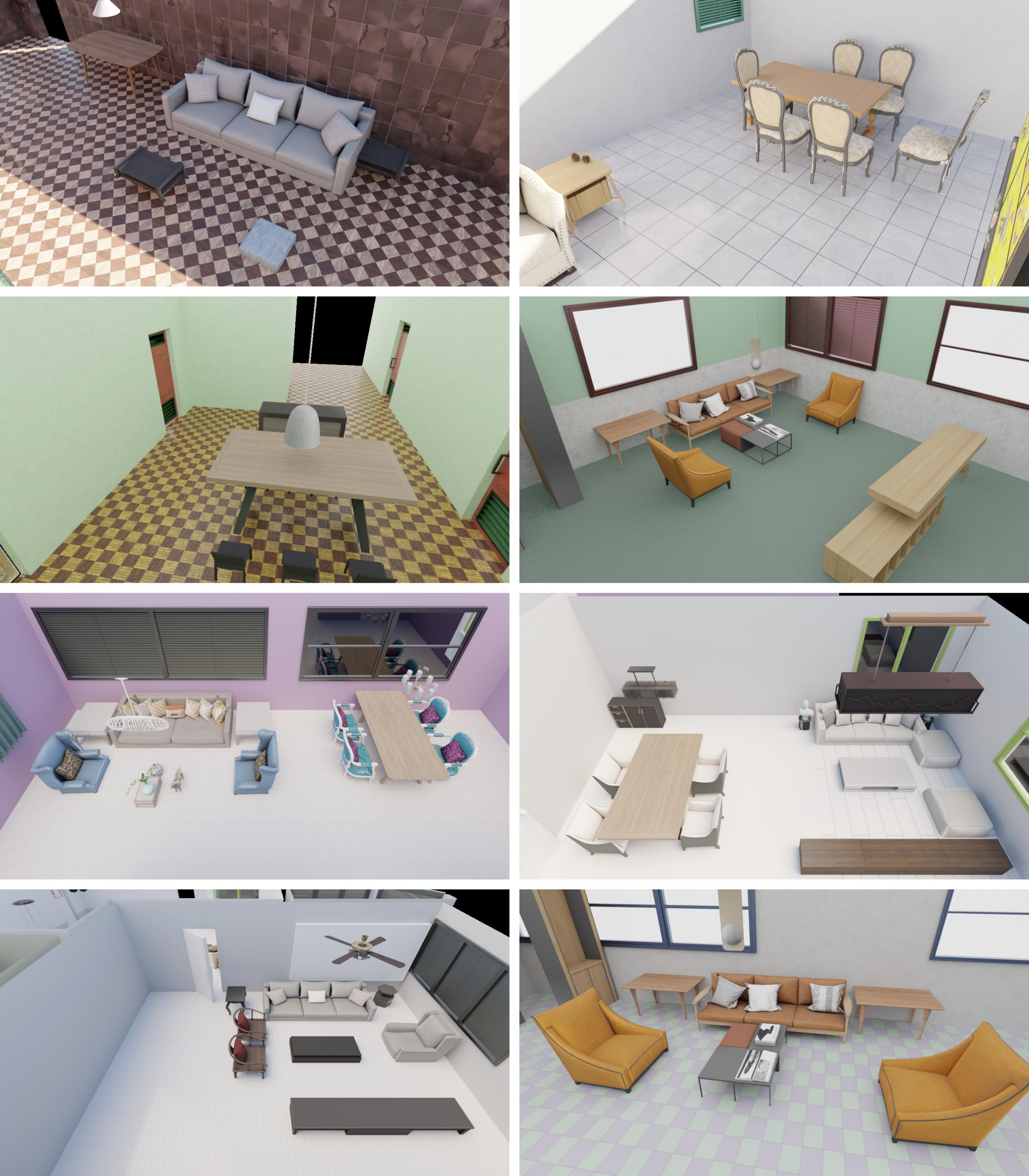}
    \caption{\textbf{Generated 3D Layouts.} Representative visualization of scenes generated by SceneFoundry. (Part 2).}
    \label{fig:render_result_2}
    \vspace{1cm}
\end{figure*}

These visualizations highlight the model's capability to handle complex spatial arrangements naturally. We observe that the generated objects are physically plausible, exhibiting proper orientations and avoiding inter-object collisions, as shown in Figure~\ref{fig:render_result_1} and Figure~\ref{fig:render_result_2}. Collectively, these qualitative results validate that our method not only adheres to rigid geometric constraints but also produces aesthetically pleasing and functionally realistic environments suitable for practical design applications.

\end{document}